\documentclass[sigconf]{acmart}
\usepackage{multirow}
\usepackage{makecell}

\AtBeginDocument{%
  \providecommand\BibTeX{{%
    \normalfont B\kern-0.5em{\scshape i\kern-0.25em b}\kern-0.8em\TeX}}}

\copyrightyear{2022}
\acmYear{2022}
\setcopyright{acmcopyright}
\acmConference[MM '22] {Proceedings of the 30th ACM International Conference on Multimedia }{October 10--14, 2022}{Lisbon, Portugal.}
\acmBooktitle{Proceedings of the 30th ACM International Conference on Multimedia (MM '22), October 10--14, 2022, Lisbon, Portugal}
\acmPrice{15.00}
\acmISBN{978-1-4503-9203-7/22/10}
\acmDOI{10.1145/3503161.3547881}

\begin{document}

\title{DSE-GAN: Dynamic Semantic Evolution Generative Adversarial Network for Text-to-Image Generation}

\author{Mengqi Huang}
\affiliation{%
  \institution{University of Science and Technology of China}
  \country{China}
}
\email{huangmq@mail.ustc.edu.cn}

\author{Zhendong Mao}
\authornote{Zhendong Mao is the corresponding author.}
\affiliation{%
  \institution{University of Science and Technology of China}
  \country{China}
}
\email{zdmao@ustc.edu.cn}

\author{Penghui Wang}
\affiliation{%
  \institution{University of Science and Technology of China}
  \country{China}
}
\email{wph0213@ustc.edu.cn}

\author{Quan Wang}
\affiliation{%
  \institution{MOE Key Laboratory of Trustworthy Distributed Computing and Service, Beijing University of Posts and Telecommunications}
  \country{China}
}
\email{wangquan@bupt.edu.cn}

\author{Yongdong Zhang}
\affiliation{%
  \institution{University of Science and Technology of China}
  \institution{Institute of Artificial Intelligence, Hefei Comprehensive National Science Center}
  \country{China}
}
\email{zhyd73@ustc.edu.cn}

\renewcommand{\shortauthors}{Mengqi Huang et al.}

\begin{abstract}
  Text-to-image generation aims at generating realistic images which are semantically consistent with the given text. Previous works mainly adopt the multi-stage architecture by stacking generator-discriminator pairs to engage multiple adversarial training, where the text semantics used to provide generation guidance remain static across all stages. This work argues that text features at each stage should be adaptively re-composed conditioned on the status of the historical stage (\emph{i.e.}, historical stage's text and image features) to provide diversified and accurate semantic guidance during the coarse-to-fine generation process. We thereby propose a novel Dynamical Semantic Evolution GAN (DSE-GAN) to re-compose each stage's text features under a novel single adversarial multi-stage architecture. Specifically, we design (1) \emph{Dynamic Semantic Evolution (DSE)} module, which first aggregates historical image features to summarize the generative feedback, and then dynamically selects words required to be re-composed at each stage as well as re-composed them by dynamically enhancing or suppressing different granularity subspace's semantics. (2) \emph{Single Adversarial Multi-stage Architecture (SAMA)}, which extends the previous structure by eliminating complicated multiple adversarial training requirements and therefore allows more stages of text-image interactions, and finally facilitates the DSE module. We conduct comprehensive experiments and show that DSE-GAN achieves 7.48\% and 37.8\% relative FID improvement on two widely used benchmarks, \emph{i.e.}, CUB-200 and MSCOCO, respectively.
\end{abstract}



\maketitle

\section{Introduction}
Text-to-Image generation (T2I) bridges the gap between the two most prevalent modalities of vision and language. Compared with other kinds of inputs (\emph{e.g.}, sketches or object layouts) for generating images, text descriptions are a flexible and convenient form to represent visual concepts. Therefore, T2I has a variety of potential applications, such as art generation\cite{zhi2017pixelbrush} and computer-aided design\cite{chen2018text2shape}. The key of T2I lies in generating realistic images with rich and vivid details that are semantically consistent with the text.

The last few years have witnessed the remarkable success of Generative Adversarial Networks (GANs) \cite{goodfellow2014generative} for T2I\cite{reed2016generative}. 
Most existing approaches adopt the multi-stage architecture \cite{zhang2017stackgan,zhang2018stackgan++}, which stacks multiple generator-discriminator pairs to engage multiple adversarial training. 
To be specific, each stage’s generator refines the previous stage’s results by increasing the image resolution and adding more details stage by stage. 
Based on this architecture, some recent works focus on introducing different cross-modal fusion mechanisms between stages to better align text and image\cite{xu2018attngan,zhu2019dm,yin2019semantics,cheng2020rifegan,ruan2021dae,tao2020df}, and others model intermediate representations (\emph{e.g.}, object layout or segmentation map) to bridge text and image smoothly \cite{hinzsemantic,qiao2021r}. %
However, most existing methods share one commonality: the image features are dynamically generated at each stage, but text features stay static on the contrary.

\begin{figure}
  \centering
  \includegraphics[width=0.9\linewidth]{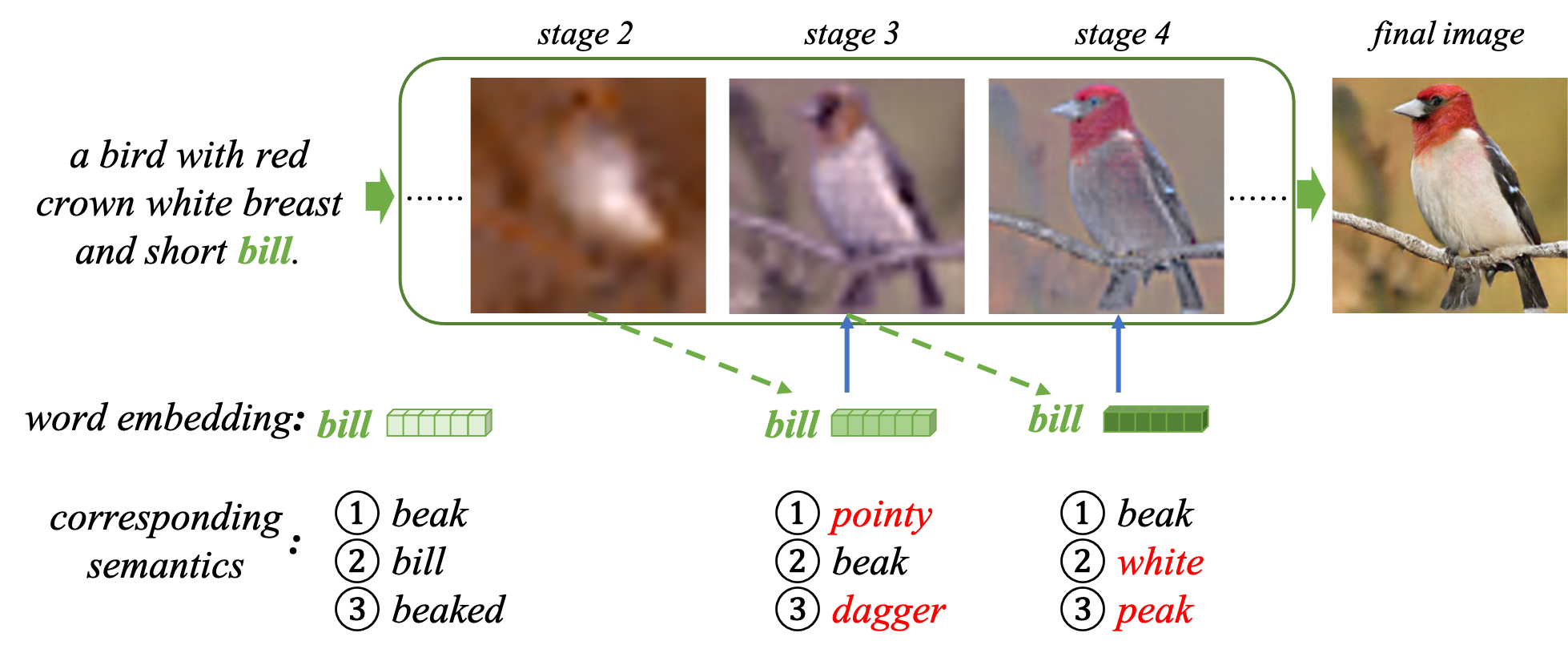}
  \caption{A real case of semantic evolution at different stages of our method in a T2I generation process. Text features consist of word embeddings, and the semantics of each word is visualized by the top-3 words with the highest cosine similar in the embedding space. Previous methods use static text features at all generation stages and the word embeddings remain unchanged, while we dynamically re-compose text features conditioned on the historical stage to provide diversified and accurate semantic guidance for each stage. Take the word ``bill'' as an example, during the text feature evolving process, new and consistent semantics (the word in red, \emph{e.g.}, ``pointy'', ``white'') are automatically activated stage by stage which finally leads to more detailed and vivid generation results. For more detailed visualization refer to Fig. \ref{interpretation}.}
  \label{intro}
\end{figure}

This paper argues that in a multi-stage T2I, the text features at each stage should be dynamically re-composed conditioned on the status of the historical stage (\emph{i.e.}, historical stage's text features and image features) to provide diversified and accurate semantic guidance. The reason is that multi-stage T2I itself is, by nature, a coarse-to-fine generation process, with the image resolution gradually increased and details gradually added stage by stage. During this process, text features should also evolve synchronously to provide semantic guidance from coarse-grained to fine-grained ({\emph{e.g.}}, from ``plant'' to ``flowers'' and then to ``red rose''), so as to better guide the image generation at each stage. Using such evolving text features brings another benefit. By dynamically re-composing text features at different stages, we are able to suppress previously used semantic information and activate new consistent ones during the generation process, which prevents the same semantics from being generated repeatedly and alleviates the repeated rendering problem. Take Fig. \ref{intro} as an example. When text features are adaptively re-composed, each word that constitutes text features evolves with new and consistent semantics stage by stage ({\emph{e.g.}}, the semantics of ``pointy'' and ``white'' gradually activated for the word ``bill''), and leads to more detailed and vivid generation results.

With this motivation, we propose a novel T2I framework dubbed as Dynamic Semantic Evolution Generative Adversarial Network (DSE-GAN), which dynamically re-composes text features conditioned on the status of the historical stage within a novel single adversarial multi-stage architecture. 
On the one hand, we devise a unique Dynamic Semantic Evolution (DSE) module, which not only dynamically selects words required to be re-composed at each stage but also dynamically re-composed them by enhancing or suppressing different granularity subspace’s semantics. 
On the other hand, to better facilitate DSE modules, we further propose a novel single adversarial multi-stage structure (SAMA), which simplifies traditional multi-stage generation by eliminating the complicated multiple adversarial training requirements and therefore allows more stages of text-image interactions. 
To be specific, as the core module, firstly DSE summarizes the generative feedback to prepare for the subsequent re-composition by predicting weights on historical image features and aggregating them into a small number of representative vectors.
Secondly, DSE calculates the correlation of each word with images' aggregated vectors to filter out words required to be re-composed through a dynamic element router, which is implemented by a gated function.
Thirdly, the selected words are further divided into multiple granularities (\emph{i.e.}, different feature dimensions) subspaces to construct a complete semantic space, where we configure a dynamic subspace router to generate the stage-aware semantic path and apply the attention mechanism to re-compose different subspaces' semantics \emph{w.r.t.} these attention scores.
In summary, we achieve a cross-modal iterative sequential method on both text and images to provide diversified and accurate semantic guidance for image generation at each stage. In this way, our method can generate images with more rich and vivid details that are semantically consistent with the text.

Our contributions can be summarized as follows: 

$\bullet$ We propose a novel sequential generation framework on both text and images for T2I, \emph{i.e.}, DSE-GAN, which dynamically re-composes text features based on the historical stage. To the best of our knowledge, this is the first framework in T2I that adaptively re-composes text features at each stage.

$\bullet$ We propose the Dynamic Semantic Evolution (DSE) module, which dynamically re-composes text features at different stages, providing diversified and accurate coarse-to-fine semantic guidance and suppressing repeated rendering.

$\bullet$ We propose a novel Single Adversarial Multi-stage Architecture (SAMA), which increases the number of stages to allow sufficient text-image interactions without additional training cost.

$\bullet$ Extensive experiments have demonstrated the superiority of our method. On CUB-200, we achieve 4.48\%, 7.48\%, and 3.12\% relative improvement on IS, FID, and R-precision, respectively. On MSCOCO, we achieve 7.36\% and a dramatic 37.8\% relative improvement on R-precision and FID, respectively. 

\begin{figure*}
  \centering
  \includegraphics[width=0.9\linewidth]{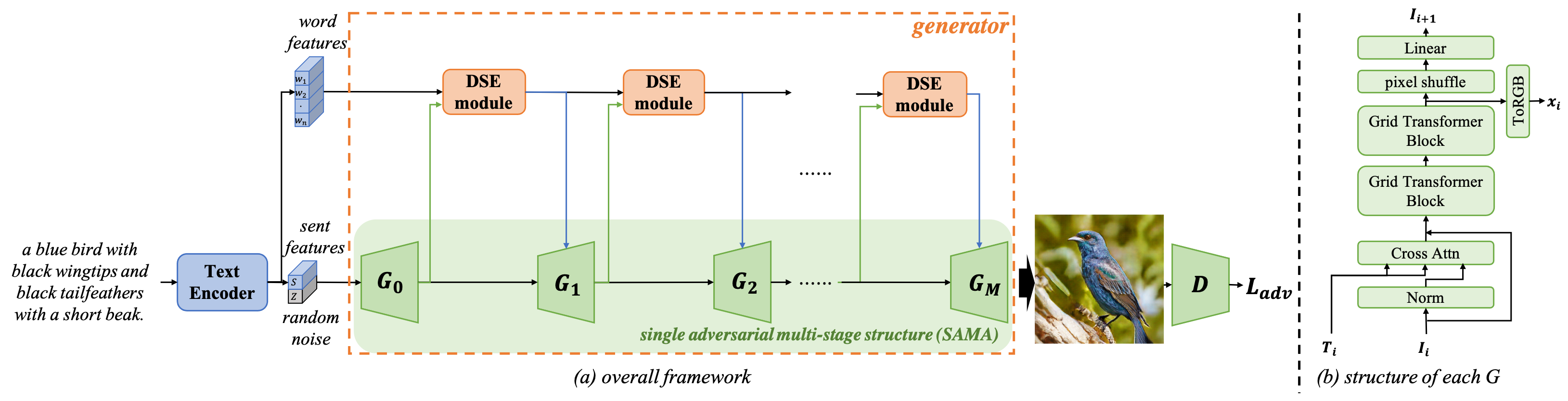}
  \caption{(a) An overview of the proposed DSE-GAN framework, which is consist of a pre-trained text encoder and a single generator-discriminator pair. The generator is consist of several Dynamic Semantic Evolution (DSE) modules and a single adversarial multi-stage structure (SAMA). Here each DSE module is responsible for re-composing the word features conditioned on the previous stage's image and word features while each sub-generator in SAMA further generates the next stage's image features under the re-composed semantic guidance. (b) The design of each sub-generator.}
  \label{fig:framework}
\end{figure*}

\section{Related Work}

\subsection{Text-to-Image Generation}
Generative Adversarial Network (GAN) \cite{goodfellow2014generative, reed2016generative} is the most popular model for T2I.
StackGAN \cite{zhang2017stackgan, zhang2018stackgan++} proposed a multi-stage generation architecture by stacking several generators as well as discriminators and engaging multiple adversarial training. Many works\cite{xu2018attngan,zhu2019dm,yin2019semantics,qiao2019mirrorgan,li2019controllable,qiao2019learn} follow this architecture, \emph{e.g.}, AttnGAN\cite{xu2018attngan} introduces the attention mechanism to generate fine-grained details at the word level.
DM-GAN\cite{zhu2019dm} adaptively refines generated images with a memory module that writes and reads text and image features. 
MirrorGAN\cite{qiao2019mirrorgan} develops a text-to-image-to-text cycle framework.
Besides, there are also many works\cite{johnson2018image, li2019object, hinz2019semantic, hinz2019generating, qiao2021r} focusing on introducing an intermediate representation (\emph{e.g.}, object bounding boxes), to smoothly bridge the text and the image. 
To avoid the cumbersome multiple adversarial training, Ming \emph{et al.} propose DFGAN\cite{tao2020df} which has only one generator-discriminator pair. 
XMC-GAN\cite{zhang2021cross} maximizing the mutual information between image and text via contrastive learning. 
Our framework also follows the single-adversarial scheme but we use more fine-grained word features and dynamically re-compose them at each stage to provide diversified and accurate semantic guidance.

Recently, auto-regressive based models such as DALL-E\cite{ramesh2021zero} and CogView\cite{ding2021cogview} shows impressive results on T2I. The general paradigm is that first using a convolutional auto-encoder to learn a codebook \cite{oord2017neural} and then applying transformer\cite{vaswani2017attention} to jointly model both text and image tokens' distribution. However, compared with GAN-based methods, the auto-regressive-based methods require a huge amount of data and resources to achieve reasonable results.

\subsection{Dynamic Network}

Dynamic neural networks have been a hot research topic in deep learning, which could adapt their structures or parameters to the given example during inference and therefore yield better representation power, adaptiveness, compatibility, and generality\cite{han2021dynamic}. 
In the literature, the research of dynamic networks can be mainly categorized into three directions, \emph{i.e.}, 
\emph{dynamic depth} for network early exiting\cite{bolukbasi2017adaptive} or layer skipping\cite{lin2017runtime,veit2018convolutional}, 
\emph{dynamic width} for skipping neurons\cite{bengio2015conditional} or channels\cite{liu2017learning}
and \emph{dynamic routing} for multi-branch or tree structure networks\cite{huang2017multi,li2020learning,yang2020resolution}.
Our method belongs in the last direction. To the best of our knowledge, the dynamic mechanism has never been studied in the field of T2I. Our model is the first one to introduce the dynamic mechanism to determine the words required to be re-composed at each stage and more importantly, re-compose the selected words' features by dynamically enhancing or suppressing different granularity subspace’s semantics.

\section{Method}

The overall framework of the proposed DSE-GAN is depicted in Fig. \ref{fig:framework}(a), which is consist of a pre-trained text encoder and a single generator-discriminator pair. The pre-trained text encoder\cite{xu2018attngan} first extracts both sentence and word features following previous works\cite{zhu2019dm,tao2020df,yin2019semantics,tan2019semantics}. The generator is consist of several Dynamic Semantic Evolution (DSE) modules and the proposed single adversarial multi-stage structure (SAMA). Here each DSE module is responsible for re-composing word features conditioned on historical image and text features while each sub-generator in SAMA further generates the next stage's image features under the re-composed semantic guidance. As for the discriminator, we adopt the one-way discriminator proposed in \cite{tao2020df} for its effectiveness and simplicity. In this section, we first elaborate on our core module, \emph{i.e.}, the DSE module, and then introduce SAMA. The objective functions for the generator as well as discriminator will be presented last.

\textbf{Notations.} Formally, $\bar{T} \in \mathbb{R}^{D_t}$ and $T_0 \in \mathbb{R}^{L_t \times D_t}$ denote original sentence and word features respectively. $L_t$ is the number of words and $D_t$ is the textual feature dimension. During the generation process, $I_i$ denotes the output image features of stage $i$ and $T_i$ denotes the word features used as semantic guidance for stage $i$. $\odot$ denotes the Hadamard Product. $A^T$ denotes the transpose of matrix $A$.

\subsection{Dynamic Semantic Evolution Module}

\begin{figure}
  \centering
  \includegraphics[width=0.9\linewidth]{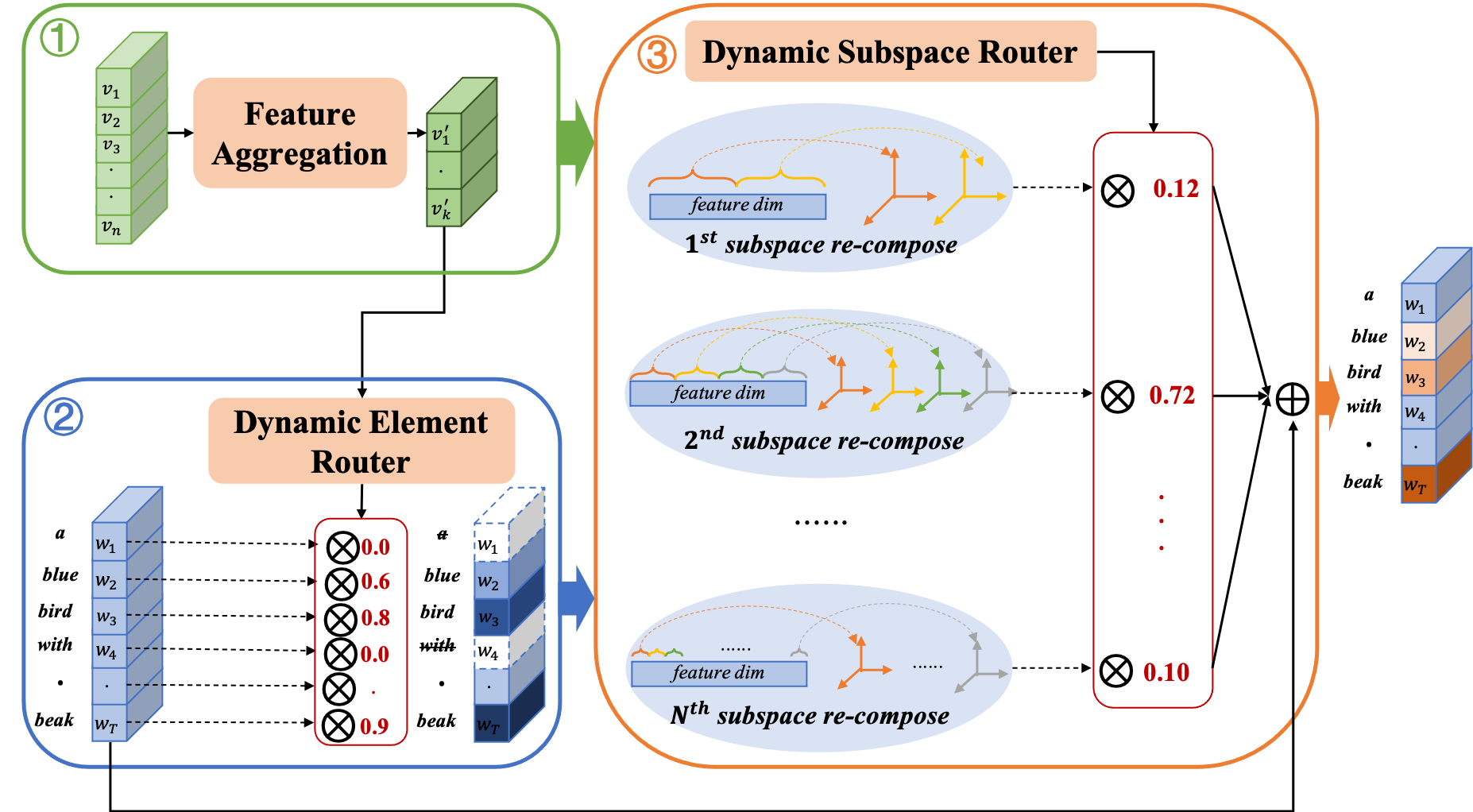}
  \caption{Illustration of the proposed DSE module, which is consist of \emph{features aggregation} sub-module for summarizing generative feedback, \emph{dynamic element router} sub-module for selecting the words required to be re-composed and \emph{dynamic subspace router} sub-module for re-composing the selected word features by dynamically enhancing or suppressing different granularity subspace’s semantics.}
  \label{fig:evolution}
\end{figure}

As shown in Fig \ref{fig:evolution}, in each generation stage $i$, the DSE module takes historical word features $T_{i-1} \in \mathbb{R}^{L_t \times D_t}$ and image features $I_{i-1} \in \mathbb{R}^{L_{i-1} \times D_{i-1}}$ as input, and outputs the re-composed word features $T_{i} \in \mathbb{R}^{L_t \times D_t}$ to provide diversified and accurate semantic guidance for current stage's image features' refinement. Here, $D_{i-1}$ is the number of channels of the image features generated by stage $i-1$ and $L_{i-1}$ is the number of image features.
The re-composition of word features is achieved by three sub-modules, \emph{i.e.}, \emph{features aggregation} sub-module for aggregating image features to summarize the generative feedback for subsequent re-composition, \emph{dynamic element routing} sub-module for dynamically selecting the words which required to be re-composed at current stage and \emph{dynamic subspace routing} sub-module for re-composing the selected word features by dynamically enhancing or suppressing different granularity subspace’s semantics. 

\textbf{Features Aggregation.} Historical image features represent the current generation status 
and are important clues for the current stage's word features' re-composition. However, we found that directly using original image features $I_{i-1}$ is neither effective nor efficient due to the modal gap and the quadratic growth of image features number $L_{i-1}$. Therefore, we first introduce 
a weighted pooling on image features to obtain a \emph{small} number of $K$ image vectors that summarizes generative feedback for subsequent re-composition. 
Concretely, we first project the image features with a learnable weight $W^a \in \mathbb{R}^{D_{i-1} \times K}$ to obtain the aggregation weight $A_{agg}$:

\begin{equation}
    A_{agg} = I_{i-1}W^a  \in \mathbb{R}^{L_{i-1} \times K},
\end{equation}
and also project the image features into the same embedding space of word features with another learnable weight $W^c \in \mathbb{R}^{D_{i-1} \times D_{t}}$:

\begin{equation}
    I_{i-1}^{'} = I_{i-1}W^c.
\end{equation}
The aggregation weight $A_{agg}$ is then normalized via a softmax operation to obtain the final results $I^{''}_{i-1}$:

\begin{equation}
    I^{''}_{i-1} = (\text{softmax}(A_{agg}))^{T}I^{'}_{i-1}  \in \mathbb{R}^{K \times D_t}.
\end{equation}
Compared with original image features $I_{i-1}$, the aggregated image features $I^{''}_{i-1}$ not only require much less computation resource since $K \ll L_{i-1}$, but also could provide better re-composition information for word features' evolution by projecting into the same embedding space
and thus both efficiency and effectiveness are achieved.

\textbf{Dynamic Element Routing.} After obtaining the aggregated image features $I^{''}_{i-1}$, we further need to select the words required to be re-composed. On the one hand, not all words are visually meaningful (\emph{e.g.,} ``I'', ``the'' and ``have'') and unnecessary emphasis on these meaningless words could negatively impact other important words' re-composition. On the other hand, different generation stages have different important words since they focus on generating different aspects of images, and therefore some important words may not need to be re-composed in a certain stage. 
Mathematically, given the previous stage's word features $T_{i-1}$ and the aggregated image features $I_{i-1}^{''}$, we first calculate the cross-modal correlation as:

\begin{equation}
    A_{cross} = T_{i-1}(I_{i-1}^{''})^{T} \in \mathbb{R}^{L_t \times K}.
\end{equation}
Then we perform mean-pooling on $A_{cross}$ to acquire the correlation of each aggregated image feature with entire word features $A_{cross}^{'}$:

\begin{equation}
    a_k^{'} = \frac{\sum_{l=0}^{L_{t}}a_{lk}}{L_{t}}.
\end{equation}
Here, $a_{lk}$ denotes the element of $l^{th}$ row and $k^{th}$ column in $A_{cross}$, and $A_{cross}^{'} = \{ a_k^{'} \in \mathbb{R}^{1} | k=0,1,...,(K-1)\} \in \mathbb{R}^{1 \times K}$. Then we apply an extra softmax operation to $A_{cross}^{'}$. The dynamic element routing is realized via a gating function to determine whether a word required to be re-composed in the next step:

\begin{equation}
    R_e = \max(0, \alpha \text{Tanh}(W_e(T_{i-1}\oplus(A_{cross}^{'}I^{''}_{i-1}))])).
\end{equation}
Here, $W_e \in \mathbb{R}^{1 \times (2\times D_t)}$ is the learnable weight. $\oplus$ denotes concatenation operation along the channel dimension. $\alpha$ is a learnable parameter to control the strength of the gating function, which we initialize as $1$. Note that if the aggregated image features number $K=1$, the term $A_{cross}^{'}I^{''}_{i-1}$ is identical to $I^{''}_{i-1}$. The output range of the dynamic element router $R_e$ is $[0, \alpha]$, and the word features are updated as $T_{i-1}^{'} = R_e T_{i-1}$, where the words' features that does not need to be re-composed will be set to $0$.

\textbf{Dynamic Subspace Routing.} Take a word feature $T_{\text{example}} \in \mathbb{R}^{D_t}$ as example, it can be divided into  different representation subspaces, \emph{i.e.}, $T_{\text{example}} \in \mathbb{R}^{h \times \frac{D_t}{h}}$, where $h$ denotes the number of subspace and $\frac{D_t}{h}$ denotes each divided subspace's dimension (\emph{i.e.}, subspace granularity). As pointed out by numbers of previous works in multi-view similarity research\cite{veit2017conditional,qu2020context,plummer2018conditional} for text-image matching and multi-head attention\cite{vaswani2017attention}, different representation subspaces contain different semantics. In this paper, instead of treating $h$ as a hyper-parameter that only re-composes word features in a fixed granularity, we first divide the word features into multiple granularities of subspaces to construct a complete semantic space and then configure a dynamic subspace router to generate stage-aware paths, which bring more accurate and diversified semantic re-composition results. Generally, given a list of subspace number $H = \{h_j | j=0,1,2,..,N-1 \}$, for each $h_j$, its corresponding divided word features $T_{i-1,j}^{'} \in \mathbb{R}^{L_t \times h_j \times \frac{D_t}{h_j}}$, and the re-composition of semantics represented by different divided subspaces is calculated via an attention mechanism as:

\begin{gather}
    \label{recompose:component}
    Q_j = T_{i-1}^{'}W_{q_j}, K_j = I_{i-1}^{''}W_{k_j}, V_j = I_{i-1}^{''}W_{v_j}, \\
    \label{recompose:output}
    O_{j} = \text{Tanh}(\text{Attention}(Q_j,K_j,V_j)) \in \mathbb{R}^{L_t \times h_j \times 1}, \\
    T_{i,j} = \text{Reverse}(T_{i-1,j}^{'} \odot O_{j}).
\end{gather}
Here, $W_{q_j}, W_{k_j} \in \mathbb{R}^{D_t \times \frac{D_t}{h_j}}, W_{v_j} \in \mathbb{R}^{D_t \times h_j}$ are learnable weights. $O_j$ is the attention output after tanh function. ``Reverse'' denotes turning the divided word features back to original size, \emph{i.e.}, $\mathbb{R}^{L_t \times h_j \times \frac{D_t}{h_j}} \rightarrow \mathbb{R}^{L_t \times D_t}$. Here, we treat the semantics of each divided subspace as a whole, \emph{i.e.}, each $\frac{D_t}{h_j}$ dimension is treated as a whole to re-compose. Since the value range of $O_j$ is $[-1, 1]$, the different semantics of the word features are suppressed or enhanced to varying degrees and realize the re-composition. To achieve the dynamic granularity subspace routing, we add another learnable weight $W_{r_j} \in \mathbb{R}^{D_t \times 1}$ to generate the routing parameters:

\begin{gather}
    \label{router:v}
    V_{r,j} = I_{i-1}^{''}W_{r_j}, \\
    \label{router:out}
    R_j = \text{Attention}(Q_j, K_j, V_{r,j}) \in \mathbb{R}^{L_t}.
\end{gather}
Therefore, the routing parameters for all granularities are represents as $\hat{R} = \{\text{mean}(R_j) | j=0,1,..N-1\} \in \mathbb{R}^{N}$. 
Different from the hard router used in previous works\cite{huang2017multi,li2020learning,yang2020resolution}, we consider a soft version via generating continuous values as different granularity subspace re-composition probabilities.
Specifically, the softmax function is applied to relax the categorical choice to a continuous and differentiable operation.
The final re-composed word features $T_i$ is computed as:
 
\begin{gather}
    \label{eq:softmax}
    \hat{R}^{'} = \text{softmax}(\hat{R}), \\
    T_i = T_{i-1} + \sum_{j=0}^{N-1} (\hat{R}^{'}_j T_{i,j}).
\end{gather}
 
\begin{figure}
  \centering
  \includegraphics[width=0.9\linewidth]{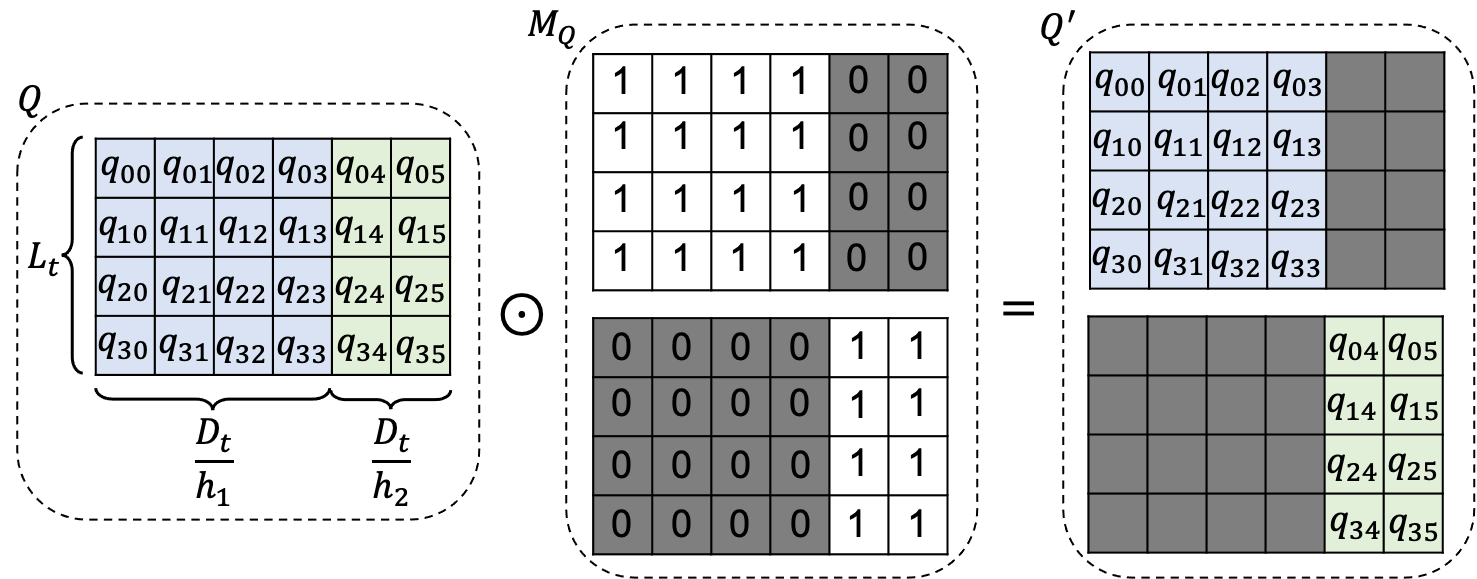}
  \caption{An example to illustrate the adjacency masks.}
  \label{fig:calculation}
\end{figure}

\textbf{Efficient implementation.} In fact, instead of doing attention calculation in each granularity respectively, we could integrate the multi-grained re-composition into single attention by placing adjacency masks to the attention and barely cost additional training time. To be specific, we replace Eq. \ref{recompose:component}, \ref{recompose:output}, \ref{router:v} and Eq. \ref{router:out} as 

\begin{gather}
    Q = T_{i-1}^{'}W_q, K = I_{i-1}^{''}W_k, V = I_{i-1}^{''}W_v, R = I_{i-1}^{''}W_r, \\
    O = \tanh(\text{Attention}(QM_Q,K,VM_V)), \\
    R = \text{Attention}(QM_Q,K,R).
\end{gather}
Here, we set the sum of different granularity subspace dimension as $D_{sum} = \sum_{j=1}^{K}\frac{D_t}{h_j}$ and the sum of different granularity subspace number as $h_{sum} = \sum_{j=1}^{K}h_j$. Therefore, $W_q, W_k \in \mathbb{R}^{D_t \times D_{sum}}, W_v \in \mathbb{R}^{D_t \times h_{sum}}, W_r \in \mathbb{R}^{D_t \times N}$ are the learnable weights. The corresponding adjacency mask for $Q$ and $V$ are $M_Q \in \mathbb{R}^{N \times L_t \times D_{sum}}, M_v \in \mathbb{R}^{N \times K \times h_{sum}}$ respectively. As illustrated in Fig. \ref{fig:calculation}, take $M_Q=\{ M_{Q,j} \in \mathbb{R}^{L_t \times D_{sum}} | j=0,...,N-1 \}$ as example, for each $M_{Q,j}$, we only set $M_{Q,j}[: , \sum_{j^{'}=0}^{j-1}\frac{D_t}{h_{j^{'}}}:\sum_{j^{'}=0}^{j}\frac{D_t}{h_{j^{'}}}]=1$ while others are set to be $0$. The binary adjacency masks $M_V$ are analogous. Therefore, different granularity re-composition will be calculated separately.

\begin{table*}
  \centering
  \begin{tabular}{ccccccc}
    \toprule
    \multirow{2}*{Methods} & \multicolumn{2}{c}{IS$\uparrow$} & \multicolumn{2}{c}{FID$\downarrow$} & \multicolumn{2}{c}{R-precision$\uparrow$(\%)} \\
    
    \cmidrule(lr){2-3} \cmidrule(lr){4-5} \cmidrule(lr){6-7}
    
    & CUB-200 & MSCOCO & CUB-200 & MSCOCO & CUB-200 & MSCOCO \\
    
    \midrule
    
    StackGAN++\cite{zhang2018stackgan++} & 4.04 $\pm$ .06 & 8.30 $\pm$ .10 & 15.30 & 81.59 & - & - \\
    AttnGAN\cite{xu2018attngan} & 4.26 $\pm$ .03 & 25.89 $\pm$ .47 & 23.98 & 35.49 & 21.65 & 55.13 \\
    DMGAN\cite{zhu2019dm} & 4.75 $\pm$ .07 & 30.49 $\pm$ .57 & 16.09 & 32.64 & 48.72 & \underline{71.08} \\
    KT-GAN\cite{tan2020kt} & 4.85 $\pm$ .04 & 31.67 $\pm$ .36 & 17.32 & 30.73 & - & - \\
    DFGAN\cite{tao2020df} & 4.86 $\pm$ .04 & 18.61 $\pm$ .12 $\dagger$  & 19.24 & 28.92 & 25.89 & 42.61 \\
    MAGAN\cite{yang2021multi} & 4.76 $\pm$ .09 & 21.66 & - & - & - & - \\
    TIME\cite{liu2021time} & \underline{4.91 $\pm$ 0.03} & 30.85 $\pm$ 0.7 & \underline{14.3} & 31.14 & - & - \\
    DAE-GAN\cite{ruan2021dae} & 4.42 $\pm$ .04 & \underline{35.08 $\pm$ 1.16} & 15.29 & 28.12 & \underline{51.64} & 70.17 \\
    XMC-GAN\cite{zhang2021cross} & - & 17.25 $\pm$ 0.04 $\ddagger$ & - & 50.08 $\ddagger$ & - & - \\
    \cmidrule(lr){1-7}
    
    R-GAN\cite{qiao2021r} & - & - & - & \underline{24.60} & - & - \\ 
    OP-GAN\cite{hinz2020semantic} & - & 27.88 $\pm$ .12 & - & 24.70 & - & 67.92 \\
    
    \cmidrule(lr){1-7}
    DALLE\cite{ramesh2021zero} & - & - & 56.10 & 27.50 & - & - \\
    Cogview\cite{ding2021cogview} & - & - & - & 27.10 & - & - \\
    \cmidrule(lr){1-7}
    
    \textbf{DSE-GAN (our method)} & \textbf{5.13 $\pm$ .03} & 26.71 $\pm$ 0.38 & \textbf{13.23} & \textbf{15.30} & \textbf{53.25} & \textbf{76.31} \\
    \bottomrule
  \end{tabular}
  \caption{Qualitative comparison between our method and other state-of-the-art. 
  $\dagger$ indicates scores computed from images generated from the open-sourced models. 
  $\ddagger$ indicates scores computed from the model trained under the same experimental setting of our methods (\emph{i.e.}, four RTX-3090 GPUs and 120 epochs on MSCOCO) using their open-source code. 
  For CUB-200, our proposed DSE-GAN achieves 4.48\%, 7.48\%, and 3.12\% relative improvement on IS, FID, and R-precision respectively. For MS-COCO, our proposed DSE-GAN achieves 37.8\% and 7.37\% relative improvement on FID and R-precision respectively.}
  \label{quantitative_results}
\end{table*}

\subsection{Dynamic Semantic Evolution Generator}

To allow more stages for text-image interactions and thereby facilitate our proposed DSE modules, we further propose a novel Single Adversarial Multi-stage Architecture (SAMA) by eliminating the complicated multiple adversarial training requirements. On the one hand, since each stage's discriminator is removed and therefore each sub-generator could not get immediate generative feedback, we adopt the grid transformer block \cite{jiang2021transgan} as our sub-generator's basic structure for its better expressivity and learning ability thanks to the built-in self-attention operations. See Fig. \ref{fig:framework}(b) for details. On the other hand, we introduce a \emph{weighted} hierarchical image integration to form the final images, which allows each sub-generator to only focus on adding corresponding details based on previous stages' generation results instead of outputting complete images. As shown in Fig.\ref{fig:framework}(a), the proposed framework has $M+1$ stages' sub-generators ($G_0, ..., G_M$), each of which take previous stage's output image features ($I_0, ..., I_{M-1}$) as input and generate images of small-to-large scales ($x_0, ..., x_M$). Especially, for the initial stage:

\begin{equation}
    I_0, x_0 = G_0(z, F^{ca}(\hat{T})),
\end{equation}
where $F^{ca}$ is the Conditioning Augmentation (CA) \cite{zhang2017stackgan} to augment the sentence feature and avoid overfitting by resampling the input sentence feature from an independent Gaussian distribution. $z$ is the random Gaussian noise vector. For the following stages:

\begin{equation}
\begin{aligned}
    & T_{i} = \text{DSE}(I_{i-1}, T_{i-1}),   i = 1, ..., M, \\
    & I_i, x_i = G_i(I_{i-1}, T_{i}), i = 1, ..., M. \\
\end{aligned}
\end{equation}  
The final target images $\mathcal{I}$ are formed by \emph{weighted} adding and sum all RGB values by skip-connection:

\begin{equation}
    \mathcal{I} = \text{sum}(\alpha_{0} x_0, \alpha_{0} x_1, ..., \alpha_{M} x_M).
\end{equation}
Here, $\{ \alpha_{i} | i=0,1,...,M \}$ are learnable weight, which are all initialize as $\frac{1}{M+1}$. During experiments, we found this small modification makes the training process more stable.

\subsection{Objective Function}

We adopt the hinge adversarial loss \cite{lim2017geometric} and apply the Matching-Aware zero-centered Gradient Penalty (MA-GP) \cite{tao2020df} to penalize the discriminator for deviating from the Nash-equilibrium. The discriminator's objective function is presented as:

\begin{eqnarray}
	\mathcal{L}^{D}=E_{x \sim p_{ {data }}}[\max (0,1-D(x, s))] \nonumber \\
	+\frac{1}{2} E_{x \sim p_{G}}[\max (0,1+D(\hat{x}, s))] \nonumber \\
	+\frac{1}{2} E_{x \sim p_{ {data }}}[\max (0,1+D(x, \hat{s}))] \nonumber \\
	+\lambda_{M A} E_{x \sim p_{ {data }}}\left[\left(\left\|\nabla_{x} D(x, s)\right\|_{2}\right.\right. 
	\left.\left.+\left\|\nabla_{s} D(x, s)\right\|_{2}\right)^{p}\right],
\end{eqnarray}
where $s$ is the matched sentence feature while $\hat{s}$ is a mismatched one. $x$ is the real image corresponding to $s$, and $\hat{x}$ is the generated image. $D(\cdot)$ is the decision given by the discriminator that whether the input image matches the input sentence. The variables $\lambda_{M A}$ and $p$ are the hyper parameters for MA-GP loss. 

The generator's adversarial objective function is presented as:

\begin{equation}
\mathcal{L}^{G}_{adv}=-E_{x \sim p_{G}}[D(\hat{x}, s)].
\end{equation}
Following \cite{xu2018attngan,zhu2019dm,tan2019semantics,qiao2019mirrorgan}, we further utilize the DAMSM loss \cite{xu2018attngan} to compute the matching degree between images and text descriptions, mathematically denoted as $\mathcal{L}_{DAMSM}$. And the Conditional Augmentation loss is defined as the Kullback-Leibler divergence between the standard Gaussian distribution and the Gaussian distribution of training text, \emph{i.e.}, 

\begin{equation}
\mathcal{L}_{C A}=D_{K L}\left(\mathcal{N}\left(\mu(\mathbf{s}), \sum(\mathbf{s})\right) \| \mathcal{N}(0, I)\right).
\end{equation}
The final objective function of the generator networks is composed of the aforementioned three terms:

\begin{equation}
    \mathcal{L}^{G} = \mathcal{L}^{G}_{adv} + \lambda_1 \mathcal{L}_{C A} + \lambda_2 \mathcal{L}_{DAMSM}.
\end{equation}

\section{Experiments}

\begin{figure*}
  \centering
  \includegraphics[width=0.9\linewidth]{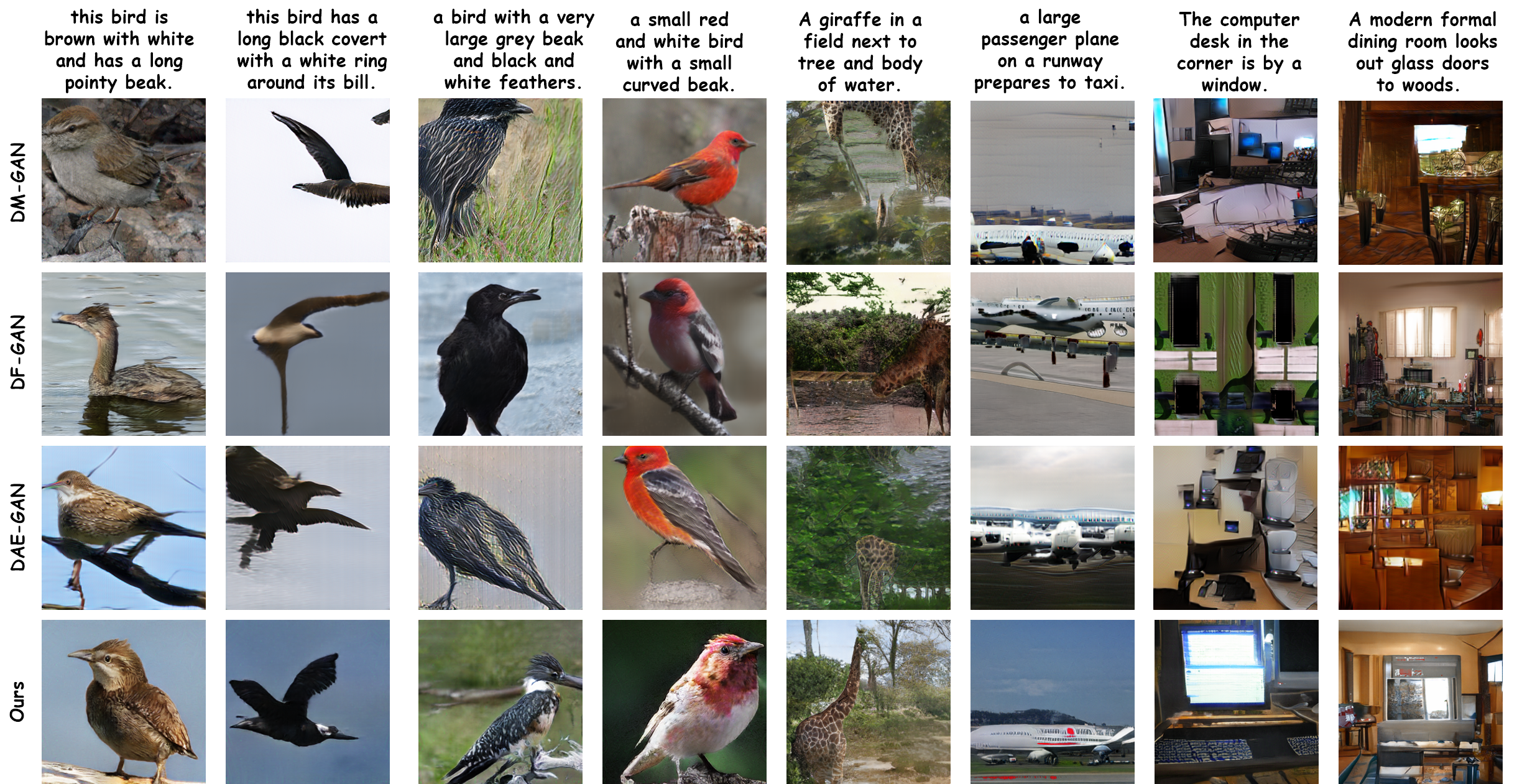}
  \caption{Examples of images synthesized by previous methods and our method on CUB-200 benchmark (1$^{st}$ - 4$^{th}$ columns) and MSCOCO benchmark (5$^{th}$ - 8$^{th}$ columns). The input text descriptions are given in the first row and the corresponding generated images from different methods are shown in the same column. Compared with previous methods, our DSE-GAN produces much more realistic results with rich and vivid details on both benchmarks. Best view in color and zoom in.}
  \label{visual}
\end{figure*}

\subsection{Experiment Settings}
\label{experiment setup}

\subsubsection{Datasets.}  
Following previous works, experiments are conducted on two benchmarks, \emph{i.e.}, the CUB-200\cite{wah2011caltech} and MSCOCO\cite{lin2014microsoft}. 

\subsubsection{Evaluation Metrics.} Following previous work, we report validation results by generating images for 30,000 random captions. 

$\bullet$ \textbf{Inception Score (IS)}, which is obtained by the pre-trained Inception-v3 network\cite{szegedy2016rethinking} to compute the KL-divergence between the conditional class distribution and the marginal class distribution. A large IS indicates better generation diversity and generation quality. \textbf{However}, as pointed out by previous work\cite{li2019object, zhang2021cross, tao2020df}, the IS completely fails in evaluating the semantic layout of the synthesized image 
and thus fails to evaluate the generation quality on MSCOCO. In this paper, we still report the IS score on MSCOCO to give a comprehensive comparison. 

$\bullet$ \textbf{Fr$\acute{\text{e}}$chet Inception Distance (FID)}, which calculates the Frechet distance between two multivariate Gaussians fit to Inception features \cite{szegedy2016rethinking} between generated and real images. The lower FID score indicates better generation quality. Compared with IS, the FID is more robust and aligns with manual evaluation.

$\bullet$ \textbf{R-precision} assesses whether the entire image matches the text description by conducting a retrieval experiment, \emph{i.e.}, generated images are used to query their corresponding text descriptions.  However, we notice that previous work computes R-precision using image-text encoders from AttnGAN \cite{xu2018attngan}, and many others use these encoders as part of their optimization function during training. This skews results: many generated models report R-precision scores significantly higher than real images. To alleviate this, we finetune CLIP \cite{radford2021learning} on both CUB-200 and MSCOCO respectively, and use the finetuned CLIP to calculate the R-precision, which is disjoint from training and better correlates with human judgments. The effectiveness of using CLIP for evaluating text-image alignment is also validated by recent studies\cite{park2021benchmark}.

\subsubsection{Implementation Details.} Our model is implemented in Pytorch \cite{paszke2019pytorch}. Adam\cite{kingma2014adam} is utilized for optimization with $\beta_1$ = 0 and $\beta_2$ = 0.99. The learning rate is set to 0.0001 for the generator and 0.0004 for the discriminator according to the Two Timescale Update Rule (TTUR) \cite{heusel2017gans}. The hyper-parameters $p$, $\lambda_{M A}$, $\lambda_1$, $\lambda_2$ are set to 6, 2, 1, 0.1 respectively. Following previous works\cite{xu2018attngan,zhu2019dm,tao2020df,hinzsemantic,tan2019semantics,qiao2019mirrorgan}, the main results are obtained trained on 600 epochs on CUB-200 and 120 epochs on MSCOCO by four RTX 3090 GPUs. 
The aggregated image number $K$ is set to be 4 and the subspace number list is set to be $\{256,128,64,32,16,8,4,2\}$ to construct a complete semantic space since the dimension of text features is $256$. 

\subsection{Comparison with state-of-the-art methods}

\subsubsection{Quantitative Results.} We compare our proposed methods with various kinds of state-of-the-art methods, including \emph{direct text-to-image}\cite{zhang2018stackgan++,xu2018attngan,zhu2019dm,tan2020kt,tao2020df,liu2021time,yang2021multi}, \emph{text-to-layout-to-image}\cite{hinz2020semantic,qiao2021r} and recent \emph{auto-regressive} methods\cite{ding2021cogview,ramesh2021zero}, on both benchmarks for all automated metrics comprehensively, as shown in Tab. \ref{quantitative_results}. For the CUB-200 benchmark, our proposed DSE-GAN outperforms previous works for all metrics, which achieves 4.48\% (from 4.91 to 5.13), 7.48\% (from 14.3 to 13.23), and 3.12\% (from 51.64\% to 53.25\%) relative improvement on IS, FID and R-precision, respectively. As for the more challenging MSCOCO benchmark, DSE-GAN dramatically improves FID from 24.60 to 15.30, a 37.8\% relative improvement over the next best model, R-GAN\cite{qiao2021r}, which requires extra object bounding box labels. DSE-GAN also outperforms others for R-precision (from 71.08 to 76.31, a 7.36\% relative improvement), which indicates the better text-image alignment of our models. Although our DSE-GAN is inferior to some previous works (\emph{e.g.}, DMGAN\cite{zhu2019dm} and DAE-GAN\cite{ruan2021dae}) for IS on MSCOCO, IS score is known for failing to evaluate the generation quality on MSCOCO \cite{zhang2021cross,li2019object,tao2020df} and the visual inspection in Fig.\ref{visual} indicates DSE-GAN’s image quality is much higher than others.

\subsubsection{Qualitative Results.} We qualitatively compare the generated images from our method with three recent state-of-the-art GAN methods, \emph{i.e.}, DMGAN\cite{zhu2019dm}, DFGAN\cite{tao2020df} and DAE-GAN\cite{ruan2021dae}. For the CUB-200 benchmark, as shown in the first four columns in Fig. \ref{visual}, our DSE-GAN generates images with many rich and vivid details which are consistent with the given textual descriptions. For example, in the 3$^{th}$ and 4$^{th}$ columns, compared with previous methods, the birds generated by our DSE-GAN have more realistic feather textures. For the MSCOCO benchmark, which has more objects and is more challenging, as shown in the last four columns of Fig. \ref{visual}, our DSE-GAN still could generate images with reasonable objects shape and different backgrounds. For example, our DSE-GAN is the only model which successfully generate ``giraffe'' in 5$^{th}$ columns, ``plane'' in 6$^{th}$ columns and the ``keyboard'', ``display screen'' in 7$^{th}$ columns. Meanwhile, our DSE-GAN also generates a much more realistic layout for the complex scene, \emph{e.g.}, the ``dining room'' with clearly ``looks out glass'' in 8$^{th}$ columns.

\begin{figure*}
  \centering
  \includegraphics[width=0.9\linewidth]{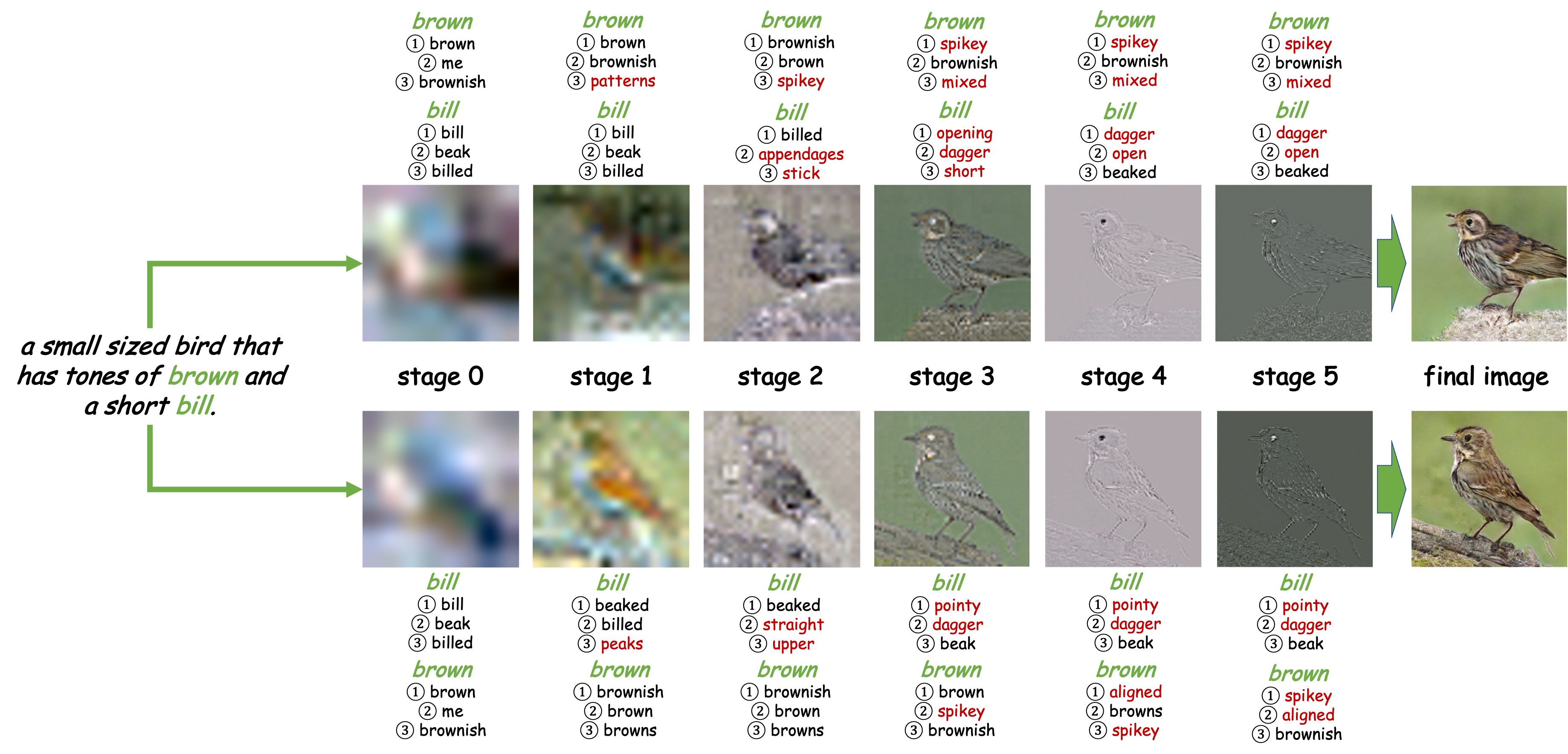}
  \caption{Two generation cases conditioned on the same text are presented to illustrate the text's semantic adaptive evolution during the generation process. We present each stage's output and the corresponding semantic visualization (top 3 most similar words in the embedding space) of ``bill'' and ``brown''. Newly activated semantics are highlighted in red. The final images are formed as the sum of all stages' output. As can be seen, the upper image flow activates ``bill'' with the semantics of ``dagger", ``open'', \emph{etc}, while the lower image flow activates ``bill'' with ``straight", ``pointy'', \emph{etc}. Both evolving semantics provide diversified and accurate guidance for each stage's generation and present different but semantically consistent appearances in the final images. The same conclusion also holds with ``brown''.}
  \label{interpretation}
\end{figure*}

\subsection{Ablations}

\begin{table}[]
    \centering
    \begin{tabular}{c c c c}
        \hline
        ID & Components & IS$\uparrow$ & FID$\downarrow$ \\
        \hline
        $\mathcal{A}$ & baseline (SAMA) & 4.84 $\pm$ .04 & 15.84 \\
        
        $\mathcal{B}$ & $\mathcal{A}$ + element routing & 4.83 $\pm$ .03 & 15.09 \\
        
        $\mathcal{C}$ & $\mathcal{A}$ + subspace routing & 4.93 $\pm$ .07 & 14.66 \\
        
        $\mathcal{D}$ & $\mathcal{A}$ + subspace routing (hard) & 4.89 $\pm$ .01 & 14.79 \\
        
        $\mathcal{E}$ & $\mathcal{A}$ + element \& subspace routing & 5.00 $\pm$ .06 & 14.02 \\
        
        $\mathcal{F}$ & \makecell[c]{$\mathcal{E}$ + image features aggregation (K=4)} & 5.09 $\pm$ .04 & 13.44 \\
        \hline
    \end{tabular}
    \caption{Effectiveness of different components.}
    \label{tab:ablation_1}
\end{table}

We thoroughly evaluate the different components of DSE-GAN and analyze their impact. All ablation results are reported on CUB-200 by two RTX-3090 GPUs. 
We verify each part of our proposed DSE module, as shown in Tab. \ref{tab:ablation_1}. Both \emph{element \& subspace routing} alone could bring improvements for our proposed SAMA baseline since the \emph{element routing} alone can be regarded as dynamically enhancing different words for each stage and the \emph{subspace routing} provides diversified as well as accurate semantic guidance for each stage. By adding the \emph{element routing}, the performance of \emph{subspace routing} is further increased, which validates the necessity of avoiding unnecessary words' re-composition for each stage. As for the \emph{subspace routing}, we also investigate the hard version router \cite{huang2017multi,li2020learning,yang2020resolution}, \emph{i.e.}, replacing softmax in Eq. \ref{eq:softmax} with Gumbel softmax, which we denotes as \emph{subspace routing (hard)}. We found that the soft version subspace routing is better than the hard version one, and we hypothesize that the reason lies in that the soft router could combine different granularity subspaces re-composition results and become more diversified and accurate. The \emph{features aggregation} sub-module further brings improvement since it summarizes the generative feedback. Compared with directly using original image features, the aggregated ones provide better re-composition information for word features’ evolution by projecting into the same embedding space as well as capturing the high-order statistics of image features, which is both effective and efficient.

\subsection{Visualization and Case Study}

To understand the text semantic adaptive evolution process, we present two generation cases conditioned on the same text, as shown in Fig. \ref{interpretation}. 
We present each stage's output and the corresponding semantic visualization (top 3 most similar words in the embedding space) of ``bill'' and ``brown''. 
Newly activated semantics are highlighted in red. The final images are formed as the sum of all stages' output. 
As can be seen, the upper image flow activates ``bill'' with the semantics of ``dagger'', ``open'', \emph{etc}, while the lower image flow activates ``bill'' with ``straight'', ``pointy'', \emph{etc}. 
Both evolving semantics provide diversified and accurate guidance for each stage's generation and present different but semantically consistent appearances in the final images. 
Meanwhile, we could observe that each important word is also dynamically determined by the historical stage whether it is required to be re-composed, \emph{e.g.}, the word ``bill'' is not re-composed at stage 1 since the generative summary of stage 0 is too coarse to infer any information for adding details for rendering ``bill''.
The same conclusion also holds with ``browns''.

\section{Conclusion}

In this paper, we propose a novel Dynamic Semantic Evolution Generative Adversarial Network (DSE-GAN) for text-to-image generation. Different from the previous methods that utilize static text features across all generation stages, our method dynamically re-composes text features at different stages conditioned on the status of the historical stage by first aggregating historical image features to summarize the generative feedback, and then dynamically selecting the words required to be re-composed at each stage as well as re-compose them by dynamically enhancing or suppressing different granularity subspaces' semantics via the novel Dynamic Semantic Evolution (DSE) modules. Moreover, the novel single adversarial multi-stage architecture (SAMA) further facilitates DSE by eliminating the complicated multiple adversarial training requirements and therefore allows more stages of text-image interactions. Comprehensive experiments on two widely used benchmarks demonstrate the superiority of our DSE-GAN.

\section{Acknowledgments}
This work is supported in part by National Natural Science
Foundation of China under Grants 62222212, U19A2057, and 61876223, Science Fund for Creative Research Groups under Grant 62121002, and
Fundamental Research Funds for the Central Universities
under Grants WK3480000008 and WK3480000010.

\bibliographystyle{ACM-Reference-Format}
\bibliography{main_paper_reference}

\end{document}